\documentclass[runningheads]{llncs}
\usepackage[T1]{fontenc}
\def\doi#1{\href{https://doi.org/\detokenize{#1}}{\url{https://doi.org/\detokenize{#1}}}}
\usepackage{graphicx}
\usepackage{tabularx}
\usepackage{multirow}
\usepackage{listings}
\usepackage{amsmath}
\DeclareMathOperator{\cosim}{cosim}

\begin{document}
\title{Can segmentation models be trained with fully synthetically generated data? }

\titlerunning{A generative model of shape and style}

\authorrunning{Fernandez et al.}

\author{Virginia Fernandez\inst{1}\orcidID{0000-0001-5984-197X} \and
Walter Hugo Lopez Pinaya \inst{1}\orcidID{0000-0003-3739-1087}  \and Pedro Borges \inst{1}\orcidID{0000-0001-5357-1673} \and Petru-Daniel Tudosiu \inst{1} \orcidID{0000-0001-6435-5079} \and Mark S. Graham \inst{1}\orcidID{0000-0002-4170-1095} \and Tom Vercauteren \inst{1}\orcidID{0000-0003-1794-0456} \and M. Jorge Cardoso \inst{1}\orcidID{0000-0003-1284-2558}}


\institute{King's College London, London WC2R 2LS, UK}

\maketitle  
\begin{abstract}
In order to achieve good performance and generalisability, medical image segmentation models should be trained on sizeable datasets with sufficient variability. Due to ethics and governance restrictions, and the costs associated with labelling data, scientific development is often stifled, with models trained and tested on limited data. Data augmentation is often used to artificially increase the variability in the data distribution and improve model generalisability. Recent works have explored deep generative models for image synthesis, as such an approach would enable the generation of an effectively infinite amount of varied data, addressing the generalisability and data access problems. However, many proposed solutions limit the user's control over what is generated. In this work, we propose brainSPADE, a model which combines a synthetic diffusion-based label generator with a semantic image generator. Our model can produce fully synthetic brain labels on-demand, with or without pathology of interest, and then generate a corresponding MRI image of an arbitrary guided style. Experiments show that brainSPADE synthetic data can be used to train segmentation models with performance comparable to that of models trained on real data.  
\end{abstract}

\section{Background}


In recent years, there has been a growing interest in applying deep learning models to medical image segmentation. Indeed, Convolutional Neural Networks are a good surrogate for manual segmentation \cite{Ronneberger2015}, which is time-consuming and requires anatomical and radiological expertise. However, deep learning models typically require large and heterogeneous training data to achieve good and generalisable results \cite{goodfellow_dl}. Yet, the access to sizeable medical imaging datasets is limited. Not only do they require specialised and costly equipment to acquire, but they are also subject to strict regulations, reduced accessibility, and complex maintenance in terms of data curation \cite{rieke_federated_learning}. Even when these datasets are accessible, labels are often scarce and task-specific, increasing the domain shift between datasets. A model which can generate images and associated labels with arbitrary contrasts and pathologies would democratise medical image segmentation research and improve model accuracy and generalisability.

Brain magnetic resonance imaging (MRI) datasets are heterogeneous as they tend to arise from a diversity of image acquisition protocols, and are partially labelled. As certain pathologies tend to be more perceptible in some MRI contrasts than others, different acquisition protocols are often followed depending on the nature of the study~\cite{mahmoud}. Furthermore, there is a significant lack of label consistency across datasets: namely, the annotated regions in any given dataset will be tailored to the study for which they were acquired \cite{Antonelli2022}.

To address the lack of comprehensive labelled data for many brain MRI segmentation tasks, domain adaptation (DA) and multi-task learning techniques can be used to create models that are robust to small or incomplete datasets \cite{Dorent2021,syntseg}. Another approach is to augment the data by applying simple transformations on individual images or by modelling the data distribution along relevant directions of variability \cite{acero}. Generative modelling, typically using unsupervised learning methods \cite{goodfellow_gans}, yields a representation of the input data distribution that does not require the user to have substantial prior knowledge about it. Some of these DL models are stochastic, allowing for continuous sampling of varied data, making them suitable for data augmentation. Such is the case for Generative Adversarial Networks (GANs) \cite{goodfellow_gans} and Variational Auto-Encoders (VAEs) \cite{kingma_vae}. In addition, some generative models allow for conditioning \cite{vqvae,spade,genmodel2016,stylebank,SEAN,resail}, opening the door to models that provide data as a function of the user's query. 

Conditional generative models have been previously applied to augment data for brain MRI segmentation tasks \cite{redgan}, but they require non-synthetic input segmentation maps. To this end, we propose brainSPADE, a fully synthetic model of the neurotypical and diseased human brain, capable of generating unlimited paired data samples to train models for the segmentation of healthy regions and pathologies. Our model, brainSPADE is comprised of two sub-models: 1) a synthetic label generator; 2) a semantic image generator conditioned on the labels arising from the label generator. Our image generator provides the user with control over the content and contrast of the output images independently. We show that segmentation models trained on the fully synthetic data produced by the proposed generative model do not only generalise well to real data but also generalise to out-of-domain distributions.

\section{Materials and Methods}

\subsection{Materials} \label{data_used}

\textbf{Data:} We trained our models on T1-weighted and FLAIR MRI images from several datasets, which all have been aligned to the MNI152-T1 template: 
\begin{itemize}
\item Training of label generator: We used quality-controlled semi-automatically generated labels from a subset of 200 patients from the Southall and Brent Revisited cohort (SABRE) \cite{SABRE}, and from a set of 128 patients from the Brain Tumour Segmentation Challenge (BRATS) \cite{Menze2015}. Healthy labels were obtained using \cite{GIF}, and are in the form of partial volume (PV) maps of five anatomical regions: cerebrospinal fluid (CSF), white matter (WM), grey matter (GM), deep grey matter and brainstem. Tumour labels were provided with BRATS. 
\item Training the image generator: we used the images and associated labels from the same SABRE and BRATS datasets, plus a subset of 38 volumes from the Alzheimer's Disease Neuroimaging Initiative 2 (ADNI2) \cite{ADNI}. 
\item Validation experiments: a hold-out set of the SABRE and BRATS datasets were used for validation, plus a subset of 30 FLAIR volumes from the Open Access Series of Imaging Studies (OASIS) \cite{LaMontagne2019} and 34 T1 volumes from the Autism Brain Imaging Data Exchange (ABIDE) dataset \cite{Martino2013}. The labelling mechanism for all datasets was the same as for the training process  \cite{GIF} except for the BRATS tumour labels, where we used the challenge manual segmentations, and the ABIDE dataset, where the CSF, GM and WM labels were generated with SPM12 (version r7771, running on MATLAB R2019a).
\end{itemize}

\noindent\textbf{Slicing process: } The proposed model works in 2D (192$\times$256). For the healthy label generator, 7008 random label slices were sampled from SABRE, and for the lesion generator, 8636 label slices (2/3 containing at least 20 lesion tumour pixels, 1/3 containing none at all) were sampled from BRATS. For the image generator, 2765 random label slices and their equivalent multi-contrast images were sampled from SABRE, ADNI2, and BRATS. Only slices containing at least 10\% of brain pixels were considered, leaving out the upper and lowermost slices. 

The code for this work was written in PyTorch (1.10.2) and will be released upon publication. The networks were trained using an NVIDIA Quadro RTX 8000 GPU and an NVIDIA DGX SuperPOD cluster.

\subsection{Methods}
\begin{figure}[t!]
\centering
\includegraphics[width=0.86\textwidth]{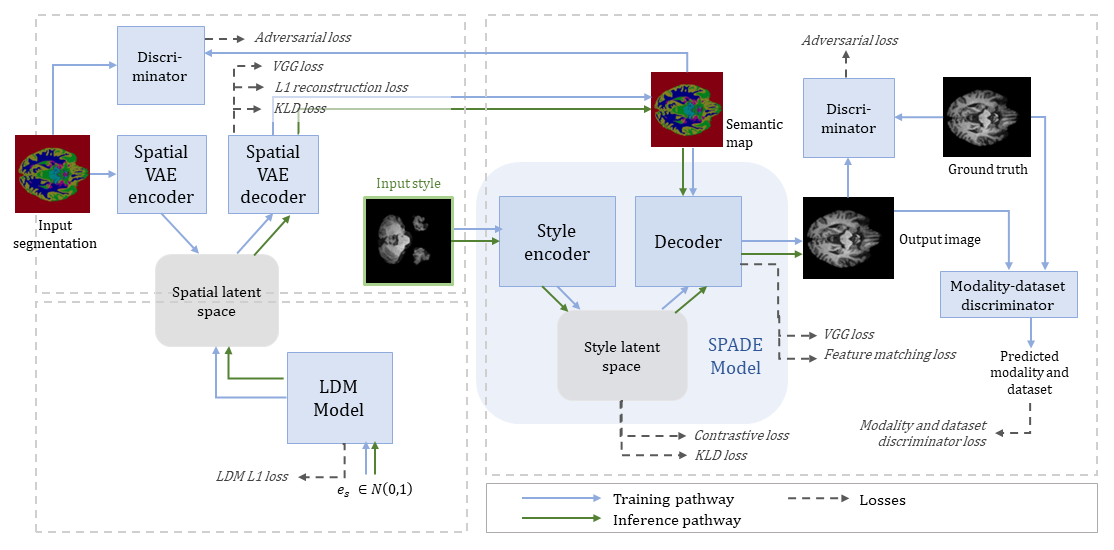}
\caption{brainSPADE pipeline: Random noise $e \in \mathcal{N}(0, \mathbf{I})$ is sampled by the LDM and incrementally denoised into a latent representation that the VAE decodes into a synthetic label. An image of the desired style is fed through the SPADE encoder, producing a latent style that, in combination with the semantic label, is decoded into the final image. Training losses have been annotated according to where they are used. Each dashed rectangle represents a different training stage.} \label{pipeline}
\end{figure}

The full model, comprising label and image generators, is depicted in Fig. \ref{pipeline}.
\\

\noindent\textbf{Label Generator:} \label{lab_gen}

Segmentations contain the morphological characteristics of the patient,
thus constituting Protected Health Information (PHI) \cite{brainprint} and requiring patient consent for sharing \cite{GDPR}. The development of a generative model of segmentations can, however, mitigate these. Segmentations are rich in phenotype information, but they lack local textures, making them challenging for standard generative models like GANs, by aggravating their considerable training instability.

To address the intrinsic limitations of GANs for label map generation, we chose to apply state-of-the-art latent diffusion models (LDMs), a generative model that samples noise from a Gaussian distribution and denoises it via a Markov chain process \cite{rombach} \cite{denoising_ho}. Coupled with a VAE, LDMs can become efficient and reliable generative models by performing the denoising process in the latent space.
Based on \cite{rombach}, first, we train a spatial VAE with two downsamplings and a latent space dimension of $48\times64\times3$, optimising the loss $L_{VAE} = D_{KLD}(E(l) \| \mathcal{N}(0, \mathbf{I})) + L_{perc.}(l, \hat{l}) + L_{adv}(D(l), D(\hat{l}))+L_{l1}(\hat{l}, l)$, where $l$ is the input ground truth segmentation maps, $\hat{l}$ is the reconstructed probabilistic partial volume segmentation map, $KLD$ is the Kullback-Leibler Distance (KLD)~\cite{kingma_vae}, $L_{perc}$ is a perceptual loss, $L_{adv}$ is an adversarial loss, computed using D, a patch-based discriminator based on \cite{taming_transformers} and $L_{l1}$ is the L1-norm. 

As a spatial VAE latent representation has a semantic context, it cannot be generated by sampling from a Gaussian distribution. Thus, an instance from that latent representation is sampled and denoised via the LDM model and then decoded with the VAE. The LDM model is based on a time-conditioned U-Net \cite{rombach} with 1000 time steps. Similar to \cite{denoising_ho}, we use a fixed variance and the reparametrized approach that predicts the added noise at each time step $\epsilon$. An L1-loss between the added noise and the predicted added noise was used to optimise the model. Two different LDM models were trained on healthy and tumour-affected semantic label slices.

\noindent\textbf{Image generator:}
SPADE \cite{spade} is an image synthesis model that generates high-quality images from semantic maps. The network is a VAE-GAN, in which the encoder yields a latent space representation of an input style image conveying the desired style, which is then used by the decoder, along with the semantic map to create an output image. The semantic maps are fed via special normalisation blocks that imprint the desired content on the output at different upsampling levels. 
We trained SPADE using the original losses from \cite{spade}: an adversarial Hinge loss based on a Patch-GAN discriminator, a perceptual loss, a KLD loss, and a regulariser feature matching loss. The ground truths for the losses corresponded to images matching the content of the input semantic map and input style image. The weights of these losses were tuned empirically. 

While the original SPADE model was found to produce high-quality outputs, the following limitations were identified:
\begin{enumerate}
    \item The latent space encoding the styles is solely optimised with the KLD loss, with no specific clustering enforcement. In our scenario, the style images are MRI contrasts that link the appearance of tissue to its magnetic properties, and thus one needs to ensure that the latent space is clustered based on the contrast, and not on aspects such as the slice number. \label{lim_1}
    \item SPADE is designed to accept categorical segmentations. However, previous work on MRI synthesis \cite{rusak} shows that partial volume maps, that associate probability of belonging to each class to each pixel, result in finer details on the output images. \label{lim_2}
    \item As explained previously, SPADE is designed to handle the style and content at different stages of the network. However, the original training process uses paired semantic maps and images to calculate the losses, making it impossible to rule out that the style latent space does not hold some information about the content of the output image. \label{lim_3}
\end{enumerate}

To address limitation \ref{lim_1}, we added two losses. First, a modality and dataset discrimination loss $L_{mod, dat}$, calculated by forward passing the generated images through a modality and dataset discriminator $D_{mod-dat}$ pre-trained on real data:
    \begin{equation}
        L_{mod,dat} = \alpha_{mod} \times BCE(mod_{\hat{i}}, mod_{i}) + \alpha_{dat} \times BCE(dat_{\hat{i}}, dat_{i});
    \end{equation}
\noindent where BCE is the binary cross entropy loss, $mod$ and $dat$ the modality and dataset (SABRE, ADNI2 etc.) predicted by $D_{mod-dat}$, $\hat{i}$ is the generated image and $i$ its equivalent ground truth. We varied $\alpha_{mod}$ and $\alpha_{dat}$ across the training, always keeping $\alpha_{mod} > \alpha_{dat}$. Secondly, a contrastive learning loss on the latent space based on \cite{contrastive_loss} is introduced as $L_{contrastive} = \cosim(E_{s}(i), E_{s}(T(i)))$, where $\cosim$ is the cosine similarity index, $E_{s}$ is the style encoder, $i$ the input style image and $T$ a random affine transformation implemented with MONAI \cite{monai}.  

Limitation \ref{lim_2} is addressed by replacing categorical labels by  probabilistic partial volume maps, which we also used to train our label generator (see \ref{lab_gen}). 

Finally, to address limitation \ref{lim_3}, we enforce the separation of the style and content generation pipelines by using different brain slices from the same volume as the style image and semantic map during training.

\begin{figure}[b!]
  \centering
  \includegraphics[width=0.85\linewidth]{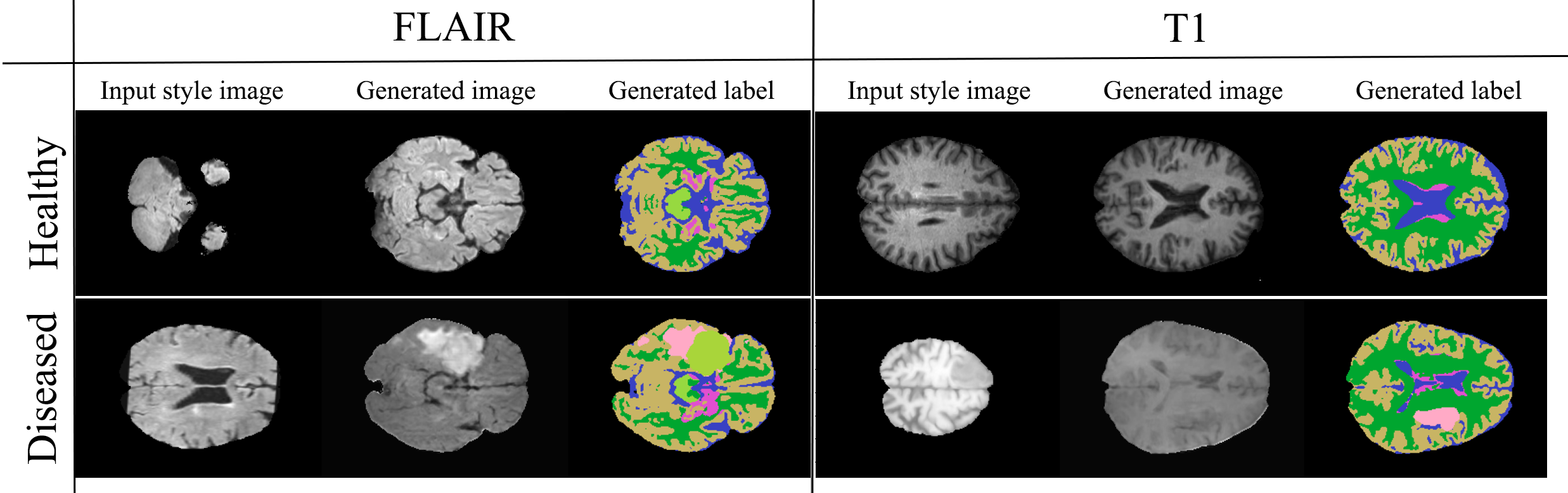} \\[\abovecaptionskip]
  \caption{Example synthetic labels and associated synthetic T1 and FLAIR images produced by our model for both our experiments, given an input style image.}\label{fig:images}
\end{figure}

Appendix \ref{app_setup} provides further information about the training process and hyperparameters. Generated neurotypical and diseased images from different modalities are displayed in Fig. \ref{fig:images}.

\subsection{Segmentation network used for the experiments}

For our segmentation experiment, we used trained instances of 2D nnU-Net \cite{nnunet} until convergence, keeping the default parameters and loss functions.

\section{Experiments}

To test whether fully synthetic datasets can be used to train segmentation models, we propose three experiments; training segmentation models on synthetic data for healthy tissue segmentation, out-of-distribution healthy tissue segmentation and tumour segmentation. All models are then tested on real data.
\subsection{Can we learn to segment healthy regions using synthetic data?}
\label{Exp1}
\subsubsection{Experiment set-up:}

In this experiment, we train two models on T1 images to segment three regions, CSF, GM and WM: 1) $R_{iod}$, trained on 7008 paired data slices from SABRE, sampled from 180 volumetric subjects; 
and 2) $S_{iod}$, trained on 20000 synthetic partial volume maps and the corresponding generated images.

\noindent The models were tested on a set of 25 holdout volumes from SABRE, and the Dice score was calculated for the volumes made up of the aggregated 2D segmentations performed by nnU-Net. 
\\
\\

\noindent \textbf{Results:} Table~\ref{dice_exp1} summarises the Dice scores per region obtained by both models on the test set. Example segmentations are depicted in Appendix \ref{app_addfig}. We compared the results using a two-sided t-test. Even though for all regions the performance of $R_{iod}$ was significantly better (p-value < 0.05), the Dice scores obtained for $S_{iod}$ were comparable to performances achieved in the literature \cite{mrbrains13}. It is important to note that the ground truth labels used for $R_{iod}$ and the ground truth labels for our test set were obtained with GIF, whereas the labels for $S_{iod}$ were obtained using the label generator, potentially creating a gap between label distributions that could explain the difference in performance. 

\begin{table}
\centering
\caption{Mean and standard deviations of Dice for each region, for experiment \ref{Exp1} (in-of-distribution (IoD)). Bold values represent statistically best performances.}\label{dice_exp1}

\newcolumntype{Y}{>{\centering\arraybackslash}c}
\newcolumntype{v}{>{\hsize=.70\hsize}Y}
\newcolumntype{s}{>{\centering\arraybackslash}X}

\begin{tabularx}{\linewidth}{|Y||s|s|}
\hline
\multirow{2}{*}{\bfseries Tissue dice}  &
\multicolumn{2}{c|}{\bfseries IoD (experiment \ref{Exp1})} \\ \cline{2-3}
& $R_{iod}$ & $S_{iod}$  \\
\hline
CSF &  \textbf{0.953$\pm$0.008} & 0.919$\pm$0.023 \\
GM &  \textbf{0.952$\pm$0.006} & 0.925$\pm$0.008 \\
WM &  \textbf{0.965$\pm$0.005} & 0.945$\pm$0.006 \\ 
\hline \end{tabularx}
\end{table}

\begin{table}
\centering
\caption{Mean and standard deviations of Dice for each region, for experiment \ref{Exp2} (out-of-distribution (OoD)). Bold values represent statistically best performances.}\label{dice_exp2}

\newcolumntype{Y}{>{\centering\arraybackslash}c}
\newcolumntype{v}{>{\hsize=.70\hsize}Y}
\newcolumntype{s}{>{\centering\arraybackslash}X}

\begin{tabularx}{\linewidth}{|s||s|s|s|s|} 
\hline
\multicolumn{5}{|c|}{\textbf{Experiment \ref{Exp2} Near out of distribution (n-OoD)}} \\
\hline
Tissue & $R_{iod}$ & $S_{iod}$ & $S_{n-ood}$ & $R_{n-ood}$ \\
\hline
CSF & 0.782$\pm$0.002 & 0.825$\pm$0.023 & 0.841$\pm$0.017 & \textbf{0.914$\pm$0.022} \\ 
GM &  0.774$\pm$0.019 & 0.881$\pm$0.008 & 0.895$\pm$0.010 & \textbf{0.971$\pm$0.011} \\ 
WM & 0.652$\pm$0.036 & 0.873$\pm$0.007 & 0.891$\pm$0.007  & \textbf{0.973$\pm$0.009} \\ 
\hline
\multicolumn{5}{|c|}{\textbf{Experiment \ref{Exp2} Far out of distribution (f-OoD)}} \\ 
\hline
Tissue & $R_{iod}$ & $S_{iod}$ & $S_{f-ood}$ & $R_{f-ood}$ \\
\hline
CSF &  0.711$\pm$0.042 & 0.736$\pm$0.054 & 0.792$\pm$0.034 & \textbf{0.830$\pm$0.050} \\ 
GM & 0.531$\pm$0.033 & 0.592$\pm$0.033 & 0.784$\pm$0.027 & \textbf{0.826$\pm$0.047} \\ 
WM & 0.447$\pm$0.180 & 0.433$\pm$0.178 & 0.809$\pm$0.031  & \textbf{0.862$\pm$0.038} \\ 

\hline \end{tabularx}
\end{table}

\subsection{Can synthetic generative models address out-of-distribution segmentation?}
\label{Exp2}
\textbf{Experiment set-up}: As an extension of Experiment \ref{Exp1}, we explored the potential of our model when it comes to handling out-of-distribution (OoD) style images, as it is able to translate between modalities and capture, to some extent, the style of unseen images. For this, we performed a near-OoD experiment (n-OoD) and a far-OoD experiment (f-OoD), using a set of slices from 25 T1 ABIDE volumes (representing n-OoD) and a set of 25 OASIS FLAIR volumes (representing f-OoD) as test target datasets. Both $R_{iod}$ and $S_{iod}$ from \ref{Exp1} were tested on these datasets. We also trained models $R_{n-ood}$ and $R_{f-ood}$ on 580 slices from five paired volumes sourced from the targets n-OoD and f-OoD distributions, serving as a reference; and models $S_{n-ood}$ and $S_{n-ood}$, on 20000 brainSPADE generated images, using the styles of unpaired images from the target n-OoD and f-OoD distributions during inference.

\noindent \textbf{Results}: The Dice scores obtained for the different structures are reported in  table~\ref{dice_exp2}. Example segmentations can be found in Appendix \ref{app_addfig}. Both $S_{iod}$ and $R_{iod}$ experienced a drop in performance when tested on n-OoD and f-OoD data, whereas $S_{n-ood}$ and $S_{f-ood}$ were significantly better (p-value < 0.0001), with dice scores closer to those achieved by models trained on paired data from the target distributions, yet significantly lower. This shows that brainSPADE has some potential for domain adaptation. 

\subsection{Can we learn to segment pathologies from synthetic data?}
\label{Exp3}

\textbf{Experiment set-up:} In this experiment, we train models on T1 and FLAIR images to segment tumours from a holdout set of sites from BRATS, unseen by brainSPADE. We trained three models: $R_{les}$ on 1064 slices from 5 paired subjects belonging to the target set; $S_{les}$ on 20000 sampled slices from brainSPADE, using the style of target T1 and FLAIR images; and $H_{les}$, combining both the training sets of $R_{les}$ and $S_{les}$. The labels for $S_{les}$ were sampled using lesion-conditioning, ensuring a balance between negative and positive samples.  The resulting models were tested on 30 test volumes from the target dataset, similarly to \ref{Exp1} and \ref{Exp2}, yielding dice scores on tumours, accuracy, precision and recall.

\noindent \textbf{Results:} The results are reported in table \ref{dice_exp3}, with visual examples available in Appendix \ref{app_addfig}. The hybrid model achieved the top performance for all metrics, having significantly better recall than $S_{les}$ and $R_{les}$.

\newcolumntype{Y}{>{\centering\arraybackslash}c}
\newcolumntype{v}{>{\hsize=.70\hsize}Y}
\newcolumntype{s}{>{\centering\arraybackslash}X}

\begin{table}[t!]
\caption{Mean and associated standard deviations of Dice, accuracy, precision and recall for the tumour segmentation task, for the four models. Bold values represent statistically best performances.}\label{dice_exp3}
\centering

\begin{tabularx}{\linewidth}{|s||s|s|s|s|} 
\hline
{\bfseries Model}  &  {\bfseries $R_{les}$ } &  {\bfseries $S_{les}$ } & {\bfseries $H_{les}$ } \\
\hline
Dice on tumour & 0.813 $\pm$ 0.174 &  0.760 $\pm$ 0.187 & 0.876 $\pm$ 0.094 \\
Accuracy & 0.995 $\pm$ 0.007  & 0.994 $\pm$ 0.006 & 0.997 $\pm$ 0.002 \\ 
Precision & 0.878 $\pm$ 0.143 & 0.864 $\pm$ 0.124 & 0.921 $\pm$ 0.061  \\
Recall  & 0.773 $\pm$ 0.209 & 0.713 $\pm$ 0.238 & \textbf{0.852 $\pm$ 0.137} \\
\hline
\end{tabularx}
\end{table}

\section{Discussion and Conclusion}

We have shown that brainSPADE, a fully-synthetic brain MRI generative model, can produce labelled datasets that can be used to train segmentation models that exhibit comparable performances to models trained using real data. The synthetic data generated by brainSPADE can not only replace real data for healthy tissue segmentation but also address pathological segmentation, as evidenced by Experiment \ref{Exp3}. In addition, because the content pathway is completely separated from the style pathway in the generative pipeline, brainSPADE makes it possible to condition on unlabelled images, producing fully labelled datasets that can help train segmentation models with reasonable performance on that target distribution. The ability to replicate, to some extent, the style of an unseen dataset is shown in Experiment 2, where using OoD images as styles for brainSPADE results in a performance boost on that dataset. 

These results open a promising pathway to tackle the lack of data in medical imaging segmentation tasks, where multi-modal synthetic data, conditioned by the user's specifications on the style and content, could not only help for data augmentation but compensate for the unavailability of paired training data. In the future, our model could be fine-tuned on more modalities and pathologies, making it generalisable and capable of addressing more complex segmentation tasks, e.g. involving small or multiple lesions.
Synthetic medical data has the advantage of not retaining any personal information on the patient, as it introduces variations on the original anatomy that should erase all traceability. Nonetheless, future work should analyse to what extent this model introduces variations on the training data and to what extent it retains it. This is key to, on the one hand, ensure that brainSPADE is stochastic and can produce an almost unlimited amount of data, and on the other hand, ensure that the training data cannot be retrieved from the model via model-inversion attacks \cite{hidano_mia}, a critical point if generative models are used as a public surrogate for real medical data. 

\pagebreak
\appendix
\section{Training set-ups} \label{app_setup}

\subsection{Training brainSPADE}

\subsubsection{Training the Label Generator}
We trained the VAE for 800 epochs using a learning rate of $5\times 10^{-5}$, Adam optimizer ($\beta_1 = 0.99, \beta_2 = 0.999$) and a batch size of 256 in an NVIDIA DGX A100 node. The training time was approximately 8 hours. 
The LDM was trained for 1500 epochs, with a learning rate of $2.5\times 10^{-5}$,  Adam Optimizer ($\beta_1 = 0.99, \beta_2 = 0.999$) and a batch size of 384 in an NVIDIA DGX A100 node. The training time was approximately 15 hours. 

\subsubsection{Training the Image Generator}
The weights that were used to balance the different losses were: $L{adv}: 1.0$, $L_{VGG}: 0.25$, $L{feature-matching}: 0.05$, $L_{KLD}: 0.001$, $L_{contrastive}: 1.0$, $L_{mod-dat}: (0.1 - 2.5; 0.05 - 0.75)$.An exponentially decaying learning rate starting at $5\times10^{-4}$ was used with an Adam Optimizer ($\beta_1 = 0.5, \beta_2 = 0.99)$ for 4800 epochs. The training time was approximately 2 weeks, using a batch size of 6 in a NVIDIA Quadro RTX 8000 GPU. 

For the training process we used the following MONAI augmentations \cite{monai}:
\begin{itemize}
    \item Random bias field augmentation, with a coefficient range of 0.2-0.6.
    \item Random contrast adjustment, with a $\gamma$ coefficient range of 0.85-1.25.
    \item Random gaussian noise addition, with $\mu$ = 0.0 and $\sigma$ range of 0.05-0.15. 
\end{itemize}

The images were normalised using Z-normalisation. 

\subsection{Training Segmentation nnU-Nets}
We used nnU-Net to perform all our segmentation experiments. nnU-Net performs automatic hyperparameter selection based on the task and input data; we downloaded the package from Github \footnote{https://github.com/MIC-DKFZ/nnUNet.git} and selected the `2d' training option. We modified the number of epochs to ensure convergence for all models.

\section{Additional figures}\label{app_addfig}
\begin{figure}
    \centering
    \includegraphics[width=0.95\textwidth]{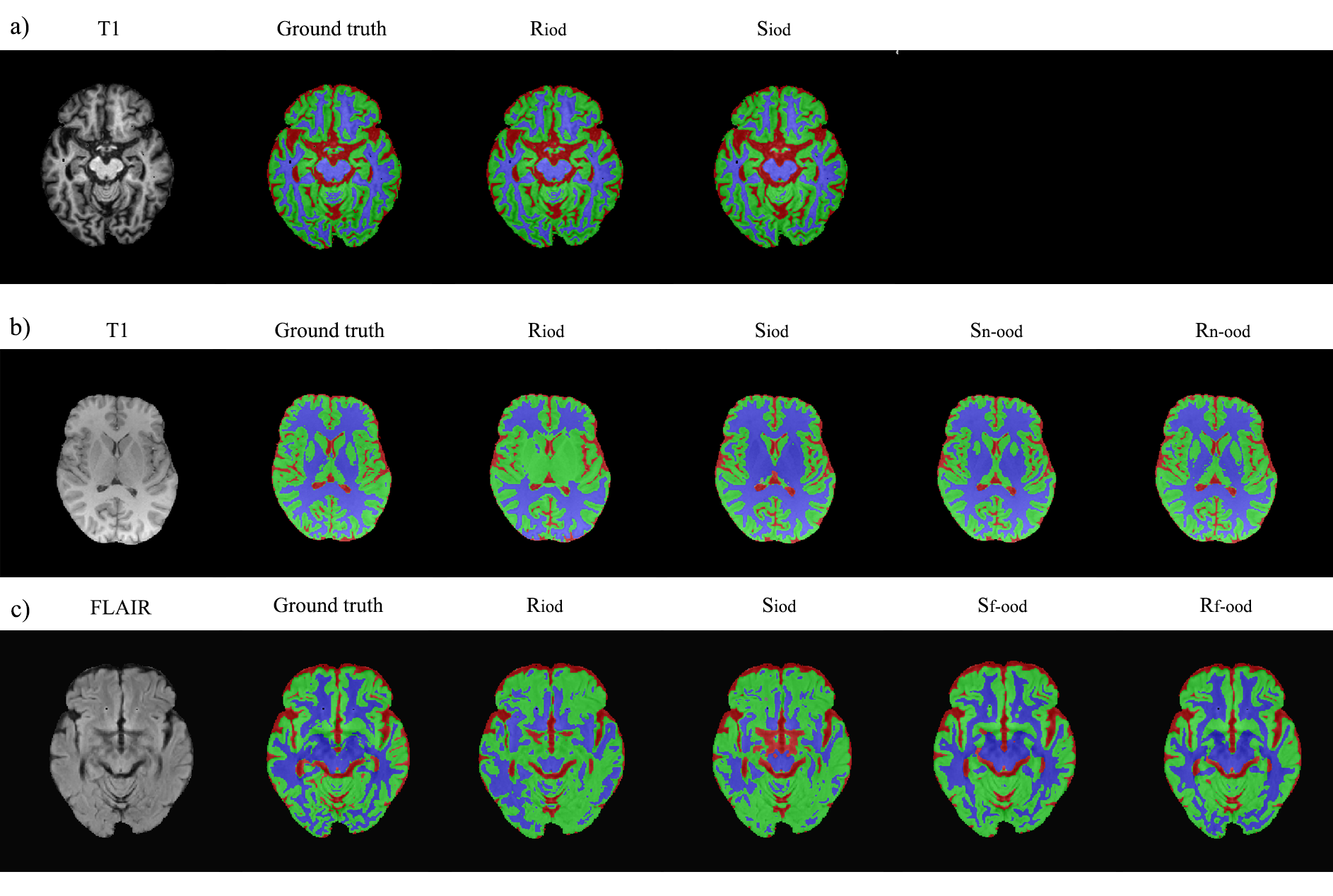}
    \caption{Example segmentations for experiments 3.1 (a) and 3.2., near-OoD (b) and far-OoD (c); the segmented regions are CSF (red), GM (green) and WM (blue).}
    \label{segm_healthy}
\end{figure}

\begin{figure}[h!]
    \centering
    \includegraphics[width=0.9\textwidth]{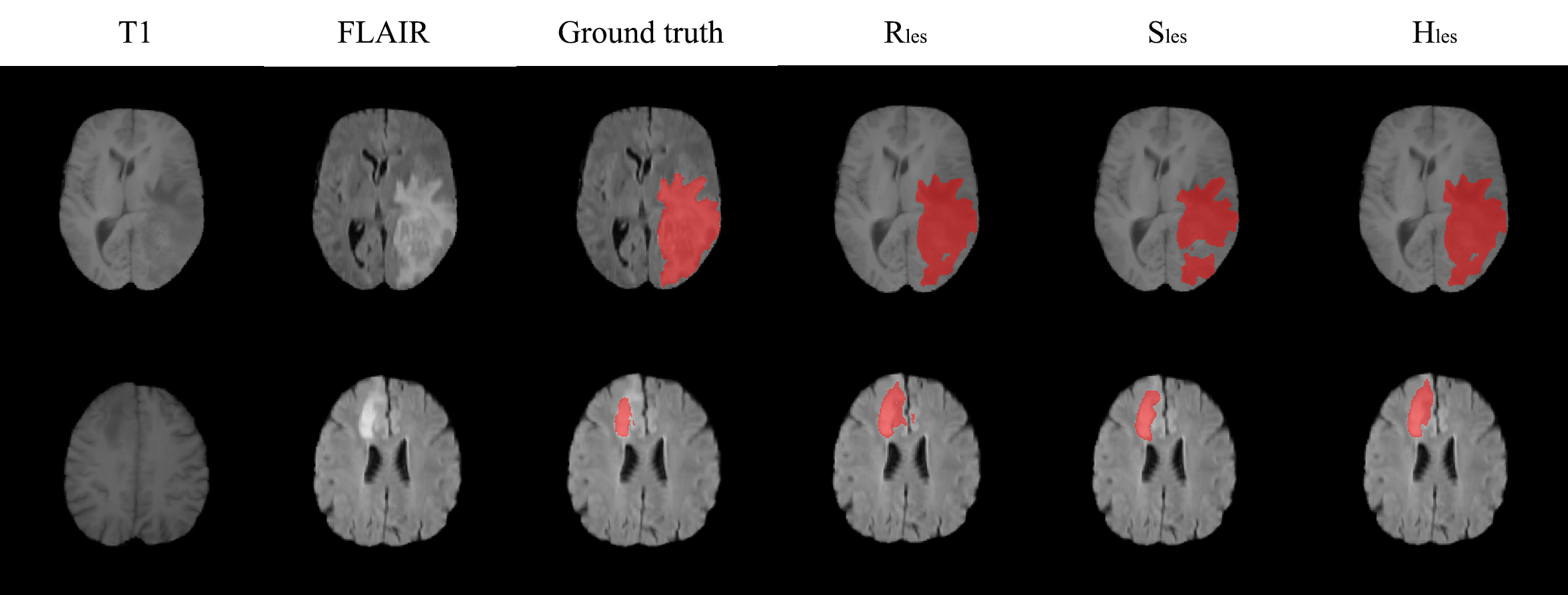}
    \caption{Example segmentations for experiment 3.3. From left to right: input T1 and FLAIR images, ground truth, and predictions of tumours made by our models $R_{les}$, $S_{les}$ and $H_{les}$, highlighted in red.}
    \label{tumours}
\end{figure}

Figures \ref{segm_healthy} and  \ref{tumours} show example segmentations from our experiments. 

\clearpage
\bibliographystyle{splncs04}
\bibliography{BIBLIOGRAPHY}

\end{document}